\documentclass[11pt,a4paper]{article}
\usepackage[hyperref]{emnlp2020}
\usepackage{times}
\usepackage{latexsym}
\usepackage{times}
\usepackage{latexsym}
\usepackage{graphicx}
\usepackage{amsmath}
\usepackage{amsthm}
\usepackage{booktabs}       
\usepackage{amsmath}
\usepackage{enumerate}
\usepackage{amsfonts}
\usepackage{algorithm}
\usepackage{algorithmic}
\usepackage{subfigure}
\usepackage{graphicx}
\usepackage{float}
\usepackage{bbm}
\usepackage{multirow}

\usepackage{colortbl}
\definecolor{gray}{RGB}{87, 87, 87}
\definecolor{red}{RGB}{173, 35, 35}
\definecolor{blue}{RGB}{42, 75, 215}
\definecolor{green}{RGB}{29, 105, 20}
\definecolor{brown}{RGB}{129, 74, 25}
\definecolor{purple}{RGB}{129, 38, 192}
\definecolor{cyan}{RGB}{41, 208, 208}
\definecolor{yellow}{RGB}{189, 167, 0}
\definecolor{Red}{rgb}{0.68, 0.05, 0.0}
\definecolor{Blue}{rgb}{0.0, 0.0, 0.61}
\definecolor{Blue1}{RGB}{214, 235, 245}
\definecolor{Blue2}{RGB}{235, 245, 250}
\definecolor{lime}{RGB}{60,179,113}
\definecolor{peach}{RGB}{255, 242, 230}
\usepackage{microtype}

\aclfinalcopy 

\title{ProphetNet: Predicting Future N-gram for Sequence-to-Sequence Pre-training}

\author{Weizhen Qi\textsuperscript{1} \thanks{ \hspace{1mm}Work is done during internship at Microsoft Research Asia.}     \thanks{ \hspace{1mm} Equal contribution} , Yu Yan\textsuperscript{2}\footnotemark[2], Yeyun Gong\textsuperscript{3}\footnotemark[2], Dayiheng Liu\textsuperscript{4}\footnotemark[2], \\
\textbf{Nan Duan\textsuperscript{3}, Jiusheng Chen\textsuperscript{2},  Ruofei Zhang\textsuperscript{2}, Ming Zhou\textsuperscript{3}}  \\
  \textsuperscript{1}University of Science and Technology of China, \textsuperscript{2}Microsoft, \textsuperscript{3}Microsoft Research Asia,
  \textsuperscript{4}Sichuan University  \\
  \texttt{\textsuperscript{1}weizhen@microsoft.com, \textsuperscript{2}\{yyua, jiuchen, bzhang\}@microsoft.com} \\  
  \texttt{\textsuperscript{3}\{yegong, nanduan, mingzhou\}@microsoft.com, \textsuperscript{4}losinuris@gmail.com}
  
  }

\date{}

\begin{document}
\maketitle
\begin{abstract}
This paper presents a new sequence-to-sequence pre-training model called ProphetNet, which introduces a novel self-supervised objective named future n-gram prediction and the proposed n-stream self-attention mechanism. Instead of optimizing one-step-ahead prediction in the traditional sequence-to-sequence model, the ProphetNet is optimized by $n$-step ahead prediction that predicts the next $n$ tokens simultaneously based on previous context tokens at each time step. The future n-gram prediction explicitly encourages the model to plan for the future tokens and prevent overfitting on strong local correlations.  We pre-train ProphetNet using a base scale dataset (16GB) and a large-scale dataset (160GB), respectively. Then we conduct experiments on CNN/DailyMail, Gigaword, and SQuAD 1.1 benchmarks for abstractive summarization and question generation tasks. Experimental results show that ProphetNet achieves new state-of-the-art results on all these datasets compared to the models using the same scale pre-training corpus.
\end{abstract}

\section{Introduction}
Large-scale pre-trained language models~\cite{devlin2018bert,radford2019language,yang2019xlnet} and sequence-to-sequence models~\cite{lewis2019bart, song2019mass, raffel2019exploring} have achieved remarkable success in downstream tasks. 

Autoregressive (AR) language modeling, which estimates the probability distribution of the text corpus, is widely used for sequence modeling and sequence-to-sequence (Seq2Seq) learning~\cite{sutskever2014sequence}. 
Recently, it also becomes one of the successful self-supervised objectives for large-scale pre-training as used in GPT-2~\cite{radford2019language}.
Specifically, given a text sequence $x = (x_1, \dots, x_T)$, AR language modeling factorizes the likelihood into a product $p(x) = \prod^{T}_{t=1} p(x_t | x_{<t})$.
In this manner, language models (LMs) and Seq2Seq models are usually trained by teacher forcing. The models are optimized to predict the next token given all previous context tokens at each time step. 

However, as discussed in previous works~\cite{pascanu2013difficulty,gulcehre2017plan,serdyuk2017twin}, AR-based models may prefer to focus on the latest tokens rather than capture long-term dependencies for the next token prediction. The reasons are as follows: (a) Local correlations such as bigram combination are usually stronger than long-term dependencies.
(b) Teacher forcing, where the model focus on one-step-ahead prediction for each time step, has no explicit bias toward future token planning and modeling. As a result, the model may learn a bias for language modeling; that is, the local token combinations' modeling is overfitting, but the global coherence and long-term dependency are underfitting~\cite{krueger2016zoneout,merity2017regularizing,serdyuk2017twin}.
During inference, the generations tend to maintain local coherence but lack meaningful global structure~\cite{li2017learning, serdyuk2017twin}, especially when we use greedy decoding instead of beam search.


In this paper, we present a new large-scale pre-trained Seq2Seq model called \textbf{ProphetNet} with a novel self-supervised objective \textbf{future n-gram prediction}.
In addition to the traditional language model (LM) or Seq2Seq model that optimizes one-step-ahead prediction, the ProphetNet also learns $n$-step ahead prediction
This future n-gram prediction is served as extra guidance that explicitly encourages the model to plan for future tokens and prevents overfitting on strong local correlations.
The hidden states of ProphetNet are forced to contain useful information for the next token and further help predict multiple future tokens.

There are two goals when designing ProphetNet: (a) the model should be able to simultaneously predict the future n-gram at each time step in an efficient way during the training phase, and (b) the model can be easily converted to predict the next token only as original Seq2Seq model for inference or fine-tuning phase.
To achieve that, we extend the two-stream self-attention proposed in XLNet~\cite{yang2019xlnet} to \textbf{n-stream self-attention}.
ProphetNet contains a main stream self-attention, which is the same as the self-attention in the original Transformer. 
Besides, we introduce $n$ extra self-attention predicting streams for future n-gram prediction, respectively. 
During training, the $i$-th predicting stream attends to the main stream's hidden states to predict the next $i$-th future token, which guarantees every $n$ continuous tokens in the target sequence are trained to predict at one time step.
Since the main stream parameters are shared with every predicting stream, we can disable the n-stream self-attention during inference. Only the next first token is predicted for each time step, which is same as the original Transformer Seq2Seq model.

For experiments, we use the proposed future n-gram prediction with the mask based auto-encoder denoising task~\cite{song2019mass,lewis2019bart} which has been proved to be effective for Seq2Seq pre-training as compared in~\citet{raffel2019exploring} for ProphetNet pre-training.  
We use two scale pre-trained datasets to pre-train ProphetNet, respectively: the base scale (16GB) dataset as used in BERT~\cite{devlin2018bert}, and the large scale (160GB) similar to BART~\cite{lewis2019bart}. 
The pre-trained ProphetNet is further fine-tuned on several NLG tasks.
Experimental results show that ProphetNet has achieved the best performance on CNN/DailyMail, Gigaword, and SQuAD 1.1 question generation tasks compared to the models using the same base scale pre-training dataset. 
For the large scale dataset pre-training experiment, ProphetNet achieves new state-of-the-art results on CNN/DailyMail and Gigaword, using only about 1/3 pre-training epochs of BART and about 1/5 pre-training corpus of T5~\cite{raffel2019exploring} and PEGASUS~\cite{zhang2019pegasus}.

\begin{figure*}[th]
    \centering
	\includegraphics[width = 5.0 in]{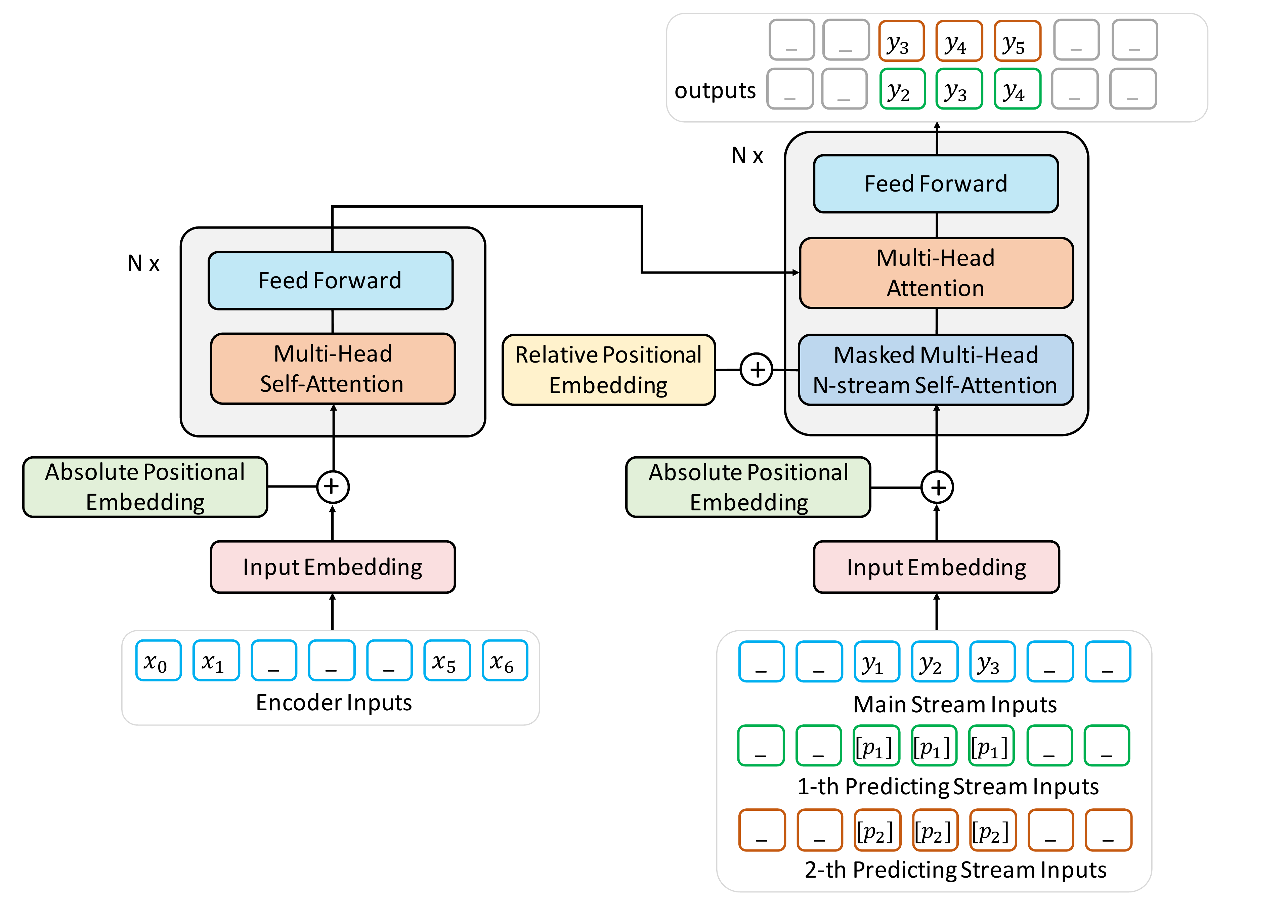}
	\caption{The architecture of ProphetNet. For simplicity, we take bigram ($n=2$) as an example to introduce ProphetNet, whose modeling target is $p(y_{t}, y_{t+1}|y_{<t}, x)$ for each time step. The left part shows the encoder of the ProphetNet which is the same as the original Transformer encoder. The right part presents the decoder of the ProphetNet which incorporates the proposed n-stream self-attention.
	For Seq2Seq pre-training, we present the example of inputs and outputs of the mask based auto-encoder denoising task. The token ``\_'' represents the mask symbol [$\mathbb{M}$]. Note that each $x_i$ and $y_i$ are the same in this task. The layer normalization and residual connection are ignored.}\label{fig:overall}
\end{figure*}

\section{ProphetNet}
We propose a new Seq2Seq pre-training model called ProphetNet, which is based on Transformer~\cite{vaswani2017attention} encoder-decoder architecture. Compared to the original Transformer Seq2Seq model, ProphetNet introduces three modifications: 
(a) The novel self-supervised objective called future n-gram prediction as described in \S~\ref{sec:m1}. 
(b) The n-stream self-attention mechanism as described in \S~\ref{sec:m2}.
(c) The mask based auto-encoder denoising task for Seq2Seq pre-training as described in \S~\ref{sec:m4}.
Figure~\ref{fig:overall} shows the architecture of ProphetNet.
Before we describe our model in detail, we first introduce the notations and sequence-to-sequence learning. 
 
\subsection{Sequence-to-Sequence Learning}
Given a text sequence pair $(x, y)$, $x = (x_1, \dots, x_M)$ is the source sequence with $M$ tokens, and $y = (y_1, \dots, y_T)$ is the target sequence with $T$ tokens.
The Seq2Seq model aims to model the conditional likelihood $p(y | x)$, which can be further factorized into a product $p(y | x) = \prod^{T}_{t=1} p(y_t | y_{<t},
x)$ according to the chain rule, where $y_{<t}$ denotes the proceeding tokens before the position $t$.
In general, the Seq2Seq model employs an encoder that aims to encode the source sequence representations and a decoder that models the conditional likelihood with the source representations and previous target tokens as inputs.
Teacher forcing is usually used for model training. The model is optimized to predict the next target token $y_t$ given the previous golden context tokens $y_{<t}$ and $x$ at each time step.

\subsection{Future N-gram Prediction} \label{sec:m1}
ProphetNet mainly changes the original Seq2Seq optimization of predicting next single token as $p(y_{t}|y_{<t}, x)$ into $p(y_{t:t+n-1}|y_{<t}, x)$ at each time step $t$, where $y_{t:t+n-1}$ denotes the next continuous $n$ future tokens.
In other words, the next $n$ future tokens are predicted simultaneously.

\begin{figure*}[h]
    \centering
	\includegraphics[width = 6.3 in]{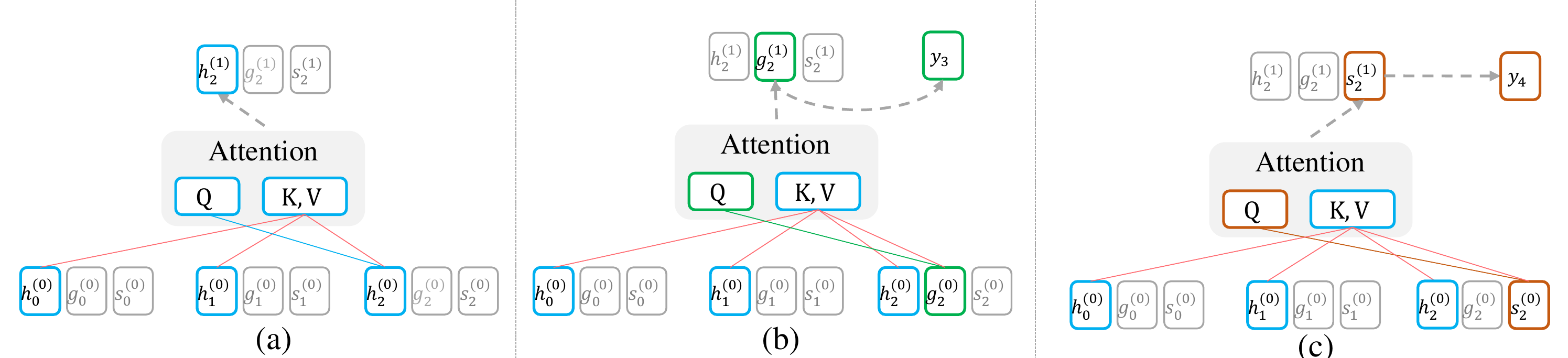}
	\caption{N-stream self-attention mechanism which contains a main stream self-attention and $n$ predicting stream self-attention. For simplicity sake, we take 2-stream self-attention ($n=2$) as an example here. Figure (a) presents the attention process of the main stream self-attention. Figure (b) and Figure (c) show the attention process of 1-st predicting stream and 2-nd predicting stream, respectively. }\label {fig:nstream}
\end{figure*}

Based on Transformer Seq2Seq architecture, ProphetNet contains a multi-layer Transformer encoder with the multi-head self-attention mechanism~\cite{vaswani2017attention} and a multi-layer Transformer decoder with the proposed multi-head n-stream self-attention mechanism.
Given a source sequence $x = (x_1, \dots, x_M)$, ProphetNet encodes the $x$ into a sequence representation, which is the same as the original Transformer encoder:
\begin{equation}
H_{\rm{enc}} = \textbf{Encoder}(x_1, \dots, x_M), 
\label{eq:encoder} 
\end{equation}
where $H_{\rm{enc}}$ denotes the source sequence representations.
On the decoder side, instead of predicting only the next token at each time step, ProphetNet decoder predicts $n$ future tokens simultaneously as we mentioned above:

\begin{align}
\small
\begin{split}
 & p(y_t|y_{<t},x), \dots, p(y_{t+n-1}|y_{<t},x)  = \textbf{Decoder}(y_{<t},H_{\rm{enc}}), \\
\end{split}
\label{decoder} 
\end{align}
where the decoder outputs $n$ probability at each time step.
The future n-gram prediction objective can be further formalized as 
\begin{align} 
    \mathcal{L} = &- \sum_{j=0}^{n-1} \alpha_j \cdot \left( \sum_{t=1}^{T-j} \log p_\theta(y_{t+j}|y_{<t},x)\right) \notag\\ 
    = &- \underbrace{\alpha_0 \cdot \left( \sum_{t=1}^T \log p_\theta(y_{t}|y_{<t},x)\right)}_{\text{language modeling loss}} \notag\\  
    &- \underbrace{\sum_{j=1}^{n-1} \alpha_{j} \cdot \left( \sum_{t=1}^{T-j} \log p_\theta(y_{t+j}|y_{<t},x)\right)}_{\text{future n-gram loss}}.
\end{align}
The above future n-gram prediction objective can be seen to consist of two parts: (a) the conditional LM loss which is the same as the original teacher forcing, and (b) the $n-1$ future token prediction losses which force the model to predict the future target tokens.
The future n-gram prediction loss explicitly encourages the model to plan for future token prediction and prevent overfitting on strong local correlations. 
$\alpha_j$ is set to balance the weights between the traditional language modeling and future n-gram prediction. 
For now we set the $\alpha_j$ with a power attenuation function as:
\begin{align}
    \alpha_{j} &= \frac{\gamma^j}{\sum_{i=0}^{n-1}\gamma^i},
\end{align}
where the $\gamma$ is the attenuation coefficient.

\subsection{N-Stream Self-Attention}~\label{sec:m2}
Ideally, we want the ProphetNet decoder to meet two requirements described in the introduction: trained to predict future n-grams simultaneously and easily disable them in inference.
In addition to the masked multi-head self-attention~\cite{vaswani2017attention} of the original transformer decoder, which is called main stream self-attention, the n-stream self-attention mechanism incorporates $n$ extra self-attention predicting streams to predict next $n$ continuous future tokens respectively at each time step.
To be concrete, the $i$-th predicting stream is responsible for modeling the probability $p(y_{t+i-1} | y_{<t}, x)$.

The n-stream self-attention mechanism is shown in Figure~\ref{fig:nstream}. In this example, $h$ stream is the main stream, $g$ stream and $s$ stream are the next 1st and 2nd token predicting stream. As shown in Figure~\ref{fig:nstream} (a), the attention mechanism of the main stream is the same as the masked multi-head self-attention in the traditional Transformer decoder, where a lower triangular matrix is set to control that each position can only attend to their previous tokens:
\begin{align}
    H^{(k+1)} = \textbf{MultiHead}(H^{(k)}, H^{(k)}, H^{(k)}),
\end{align}
here we use $H^{k}= (h^{(k)}_0, \dots, h^{(k)}_T)$ to denote the sequence of the $k$-th layer hidden state of the main stream. Implement of $\textbf{MultiHead}$ can be referenced to  Transformer~\cite{vaswani2017attention}.

The $i$-th predicting stream predicts the next $i$-th token based on the previous main stream hidden states at each time step.
In other words, the $i$-th predicting stream predicts the $y_t$ based on the previous tokens $y_{<t-i+1}$.  
In this  bigram ($n=2$) example, Figure~\ref{fig:nstream} (b) shows the  $1$-st predicting stream and its hidden state is calculated as:

\begin{small}
\begin{align}
    g^{(k+1)}_{t-1} = \textbf{MultiHead}(g^{(k)}_{t-1}, H^{(k)}_{<t} \oplus g^{(k)}_{t-1}, H^{(k)}_{<t}\oplus g^{(k)}_{t-1}),
\end{align}
\end{small}
where $g^{(k+1)}_{t-1}$ denotes the $k+1$-th layer hidden state of the $1$-st predicting stream at time step $t-1$, and $\oplus$ denotes concatenation operation. To calculate $g^{(k+1)}_{t-1}$, $g^{(k)}_{t-1}$ is taken as the attention query while the attention value and key are previous $t$ hidden states of the main stream. Besides we take $g^{(k)}_{t-1}$ as attention value and key to make the $g^{(k+1)}_{t-1}$ be position-aware. The $g^{(k+1)}_{t-1}$ is finally used to predict $y_{t}$.

Similarly, the hidden state of the $2$-nd predicting stream is calculated by:

\begin{small}
\begin{align}
    s^{(k+1)}_{t-1}= \textbf{MultiHead}(s^{(k)}_{t-1}, H^{(k)}_{<t} \oplus s^{(k)}_{t-1}, H^{(k)}_{<t}\oplus s^{(k)}_{t-1}),
\end{align}
\end{small}
where $s^{(k+1)}_{t-1}$ denotes the $k+1$-th layer hidden state of the $2$-nd predicting stream at time step $t-1$, which will be finally used to predict $y_{t+1}$.
Although the calculations of $g_{t-1}$ for $y_{t}$ prediction and $s_{t-1}$ for $y_{t+1}$ prediction are very similar, they are distinguished by different initialization tokens, absolute position embedding, and relative positional calculations.

We share the parameters of each predicting stream and main stream during training. 
Therefore, we can easily convert the ProphetNet decoder to the traditional Transformer decoder by disabling all the predicting streams during inference or fine-tuning. Besides, since each predicting stream is initialized with special tokens rather than the previous token, we combine the absolute positional embedding and T5~\cite{raffel2019exploring} proposed bucket relative positional calculation to enhance the positional information in our decoder. 

\subsection{Seq2Seq Pre-training on Denoising Task}~\label{sec:m4}
We pre-train the ProphtNet on the large-scale unlabeled text corpus with the auto-encoder denoising task widely used for Seq2Seq pre-training~\cite{song2019mass, lewis2019bart, raffel2019exploring}. 


This paper uses token span masking as our denoising task, which is the same as the MASS~\cite{song2019mass}.
As shown in Figure~\ref{fig:overall}, we mask out some token spans of the original text as the encoder input, and the model learns to recover the masked tokens.
Besides, unlike MASS learns to recover one next token at each time step, ProphetNet learns to recover the next $n$ future tokens within each masked token span.

\section{Experiments and Results}
In this section, we describe the experimental details and results.
We first describe the details of ProphetNet pre-training in \S~\ref{sec:exp1}.
Then we fine-tune the ProphetNet on two downstream NLG tasks, including text summarization as described in \S~\ref{sec:exp2} and question generation as reported in \S~\ref{sec:exp3}.
We report the experiment of large-scale pre-training in \S~\ref{sec:exp4}. Results without pre-training are compared in \S~\ref{sec:exp5}. We set predicting future gram length into 2 according to the analysis in \S~\ref{sec:exp6}.

\subsection{ProphetNet Pre-training}\label{sec:exp1}
\paragraph{Model Configuration} 
Our model is based on Transformer~\cite{vaswani2017attention} encoder-decoder structure. 
We pre-train the ProphetNet, which contains a 12-layer encoder and 12-layer decoder with 1024 embedding/hidden size and 4096 feed-forward filter size. The batch size and training steps are set to 1024 and 500K, respectively. 
We use Adam optimizer~\cite{kingma2014adam} with a learning rate of $3 \times 10^{-4}$ for pre-training. 
The implement of ProphetNet is also uploaded in the attachment.
Considering the training cost, we set the $n$ to be 2 for ProphetNet in the following experiments. Further discussions are shown in \S~\ref{sec:exp6}.

\paragraph{Pre-Training Dataset} 
Following BERT~\cite{devlin2018bert}, we use BookCorpus~\cite{zhu2015aligning} and English Wikipedia (16GB in total) to pre-train ProphetNet.
We pre-train ProphetNet on this 16GB dataset with $16 \times 32$GB NVIDIA V100 GPUs.
Note that we also pre-train ProphetNet on a larger scale dataset described in \S~\ref{sec:exp4}.

\paragraph{Pre-Training Setting}
The input length of ProphetNet is set to 512.
We randomly mask a continuous span in every 64 tokens. 
80\% of the masked tokens are replaced by [$\mathbb{M}$], 10\% replaced by random tokens, and 10\% unchanged. 
The masked length is set to 15\% of the total number of tokens. Considering the computational cost, we follow MASS~\cite{song2019mass}, where the decoder only predicts the masked fragment. The attenuation coefficient $\gamma$ is set to 1.0.

\begin{table*}[th] 
\small
\begin{center}
  \begin{tabular}{lccc}
    \toprule
 Method & ROUGE-1 & ROUGE-2 & ROUGE-L \\
 \midrule
LEAD-3~\cite{nallapati2017summarunner} &40.42 &17.62 &36.67 \\
PTGEN~\cite{see2017get}  & 36.44 & 15.66 & 33.42\\
PTGEN+Coverage~\cite{see2017get}  &39.53 &17.28 &36.38\\
S2S-ELMo~\cite{edunov2019pre}  & 41.56 & 18.94 & 38.47\\
Bottom-Up~\cite{gehrmann2018bottom}  & 41.22 & 18.68 & 38.34\\
  BERTSUMABS~\cite{liu2019text} & 41.72 & 19.39 & 38.76\\
BERTSUMEXTABS~\cite{liu2019text} & 42.13 & 19.60 &39.18 \\ 
  MASS~\cite{song2019mass} & 42.12  &19.50&  39.01\\
 UniLM~\cite{dong2019unified} & 43.33  &20.21&   40.51\\
  \hline
  ProphetNet &\textbf{43.68} & \textbf{20.64} &  \textbf{40.72}\\
 \bottomrule
\end{tabular}
\end{center}
\caption{Results on the CNN/DailyMail test set.}
\label{tab:cnn} 
\end{table*}

\subsection{Fine-tuning on Text Summarization}\label{sec:exp2}
As a typical NLG task, abstractive text summarization aims to generate a short and fluent summary of a long text document.
We fine-tune and evaluate ProphetNet on the two widely used text summarization datasets: (a) the non-anonymized version of the CNN/DailyMail dataset~\cite{see2017get}, and (b) Gigaword corpus~\cite{rush2015neural}.

\paragraph{CNN/DailyMail} 
We use Adam optimizer~\cite{kingma2014adam} with a peak learning rate $1 \times 10^{-4}$.
The batch size, warmup steps, and the total fine-tune epoch are set to 512, 1000, and 10. 
We limit the length of the output to between 45 and 110 tokens with a 1.2 length penalty during inference. 
We set beam size to 5 and remove the duplicated trigrams in beam search~\cite{fan2017controllable}. 

We compare our ProphetNet against following baselines:
\textbf{LEAD-3}~\cite{nallapati2016abstractive} which takes the first three sentences as the summary;
\textbf{PTGEN}~\cite{see2017get} which is Seq2Seq model incorporated with the pointer-generator network;
\textbf{PTGEN+Coverage}~\cite{see2017get} which introduce a coverage mechanism to PTGEN;
\textbf{Bottom-Up}~\cite{gehrmann2018bottom} which employs a bottom-up content selector based on Seq2Seq model;
\textbf{S2S-ELMo}~\cite{edunov2019pre} which uses the pre-trained ELMo~\cite{peters2018deep} representations.
Besides, we also compare our method with several pre-training based strong baselines: \textbf{BERTSUMABS}~\cite{liu2019text}, \textbf{MASS}~\cite{song2019mass}, and \textbf{UniLM}~\cite{dong2019unified}. 
These pre-training-based strong baselines are all pre-trained on the same 16GB BookCorpus + English Wikipedia dataset as ProphetNet. 

Following~\citet{see2017get}, we report the F1 scores of ROUGE-1, ROUGE-2 and ROUGE-L~\cite{lin2004rouge}.~\citet{du2017learning}
The results are presented in Table~\ref{tab:cnn}.
From the results, we can see that the ProphetNet achieves the best performances on all metrics.

\begin{table}[th] 
\small
\begin{center}
  \begin{tabular}{lcccccl} 
    \toprule
    Method & R-1 & R-2 & R-L \\
    \midrule
 OpenNMT~\cite{klein2017opennmt} & 36.73 & 17.86 & 33.68 \\
Re3Sum~\cite{cao2018retrieve} & 37.04 & 19.03 & 34.46 \\
MASS~\cite{song2019mass} & 38.73 & 19.71 & 35.96 \\
 UniLM~\cite{dong2019unified} & 38.45  & 19.45 &   35.75\\
 \hline
 ProphetNet &   \textbf{39.55}  & \textbf{20.27}   & \textbf{36.57}\\
  \bottomrule
\end{tabular}
\end{center}
\caption{Results on Gigaword test set. R is short for ROUGE.}\label{tab:gigaword}
\end{table}

\paragraph{Gigaword} 
We use Adam optimizer with a peak learning rate $1 \times 10^{-4}$.
The batch size is set to 128 and warm up steps to 1000. We fine-tune model 10 epochs with future bigram prediction training.
During inference, we set the length penalty to 1.0 and beam size to 4. We set the hyper-parameters according to the performance on the dev set.

We compare our ProphetNet against following baselines:
\textbf{OpenNMT}~\cite{klein2017opennmt} which implements the standard Seq2Seq model with attention mechanism;
\textbf{Re3Sum}~\cite{cao2018retrieve} which employs an extended Seq2Seq model to generate summaries based on the retrieved candidate summaries.
And two pre-training based strong baselines:  \textbf{MASS}~\cite{song2019mass}, and \textbf{UniLM}~\cite{dong2019unified}.
The results are presented in Table~\ref{tab:gigaword}.
It can be observed that ProphetNet outperforms previous models on all metrics.

\begin{table}[th] 
\small
\begin{center}
  \begin{tabular}{lcccccl} 
    \toprule
    Method & B4 & MTR & R-L \\
    \midrule
 CorefNQG\scriptsize{~\cite{du2018harvesting}} & 15.16 & 19.12 & - \\
 SemQG\scriptsize{~\cite{zhang2019addressing}} & 18.37 & 22.65 & 46.68 \\
 UniLM\scriptsize{~\cite{dong2019unified}} & 21.63    &25.04&   51.09\\
  ProphetNet &   \textbf{23.91}  & \textbf{26.60}   & \textbf{52.26}\\ \hline
 MP-GSN\scriptsize{~\cite{zhao2018paragraph}} & 16.38 & 20.25 & 44.48 \\
  SemQG\scriptsize{~\cite{zhang2019addressing}} & 20.76 & 24.20 & 48.91 \\
 UniLM\scriptsize{~\cite{dong2019unified}} & 23.08    &25.57&   52.03\\
 ProphetNet &   \textbf{25.80}  & \textbf{27.54}   & \textbf{53.65}\\
  \bottomrule
\end{tabular}
\end{center}
\caption{Results on SQuAD 1.1 test set (with reference of ~\citet{du2017learning} tokenized). B4 is short for BLEU-4, MTR is short for METEOR, and R-L is short for ROUGE-L. The same model is used to evaluate on the two different data splits.} \label{tab:squad-recover}
\end{table}

\subsection{Fine-tuning on Question Generation}\label{sec:exp3}
The answer-aware question generation task~\cite{zhou2017neural} aims to generate a question that asks towards the given answer span based on a given text passage or document.
We conduct experiments on this task to further evaluate the ProphetNet model.  Following~\citet{du2017learning}, we split the SQuAD 1.1~\cite{rajpurkar2016squad} dataset into training, development and test sets. We also report the results on the data split as did in~\citet{zhao2018paragraph}, which reverses the development set and test set.

The question generation task is typically formulated as a Seq2Seq problem. 
The input passage and the answer are packed as ``answer [SEP] input passage'' as input, and the question is used as the target output sequence.
We fine-tune the ProphetNet model 10 epochs in the training set and report the results of the two kinds of data splits as mentioned above. 
The first 512 tokens of the passage are fed to the model. 
The peak learning rate is $1 \times 10^{-5}$ and the batch size is set to 28.

We compare ProphetNet against the following models: \textbf{CorefNQG}~\cite{du2018harvesting} which employs a feature-rich encoder based on Seq2Seq model; \textbf{MP-GSN}~\cite{zhao2018paragraph} which incorporates a gated self-attention encoder with maxout pointer; \textbf{SemQG}~\cite{zhang2019addressing} which introduces two semantics-enhanced rewards for Seq2Seq model training.
Besides, we also compare our model with \textbf{UniLM}~\cite{dong2019unified}, which is the previous state-of-the-art on this task.

The results, according to the references provided by ~\citet{du2017learning} is shown in Table~\ref{tab:squad-recover}. The same model and inference hyper-parameters are used for the two different data split with swapped dev and test set. It can be seen that ProphetNet outperforms all previous methods with significant improvement. 

\begin{table*}[htbp] 
\small
\begin{center}
\begin{tabular}{clcccl}
\toprule
Dataset & Method & Corpus & R-1 & R-2 & R-L\\
 \midrule
 \multirow{5}{*}{CNN/DailyMail}
  & T5~\cite{raffel2019exploring} & 750GB &  43.52  & \textbf{21.55} & 40.69\\
 & PEGASUSLARGE (C4)~\cite{zhang2019pegasus} & 750GB & 43.90 & 21.20 & 40.76\\
 & PEGASUSLARGE (HugeNews)~\cite{zhang2019pegasus} & 3800GB & 44.17 &21.47 &41.11\\
 & BART~\cite{lewis2019bart}& 160GB  & 44.16 & 21.28 & 40.90\\
 &ProphetNet & 160GB & \textbf{44.20} & 21.17 & \textbf{41.30}\\
 \midrule
 \multirow{3}{*}{Gigaword} & PEGASUSLARGE (C4)~\cite{zhang2019pegasus} & 750GB &   38.75  & 19.96   & 36.14\\
 &PEGASUSLARGE (HugeNews)~\cite{zhang2019pegasus} & 3800GB & 39.12  & 19.86   & 36.24\\
  &ProphetNet & 160GB & \textbf{39.51} & \textbf{20.42} & \textbf{36.69}\\ 
 \bottomrule
\end{tabular}
\end{center}
\caption{Results on the CNN/DailyMail and Gigaword test sets of large-scale pre-training models. R is short for ROUGE, and Corpus denotes the size of the pre-training data.}\label{tab:large} 
\end{table*}

\subsection{Large-scale Pre-training}~\label{sec:exp4}


Recent works show that the pre-trained model's performance on the downstream task can be improved when using larger scaled pre-training corpora~\cite{lewis2019bart, raffel2019exploring}. 
We also pre-train ProphetNet on the 160GB English language corpora of news, books, stories, and web text, which is similar\footnote{Due to CC-News is not officially released, we use similar public news corpus REALNEWS ~\cite{zellers2019defending}} to the corpus used in BART~\cite{lewis2019bart}.
The model configuration is the same as described in \S~\ref{sec:exp1}.
We fine-tune the ProphetNet on two downstream tasks CNN/DailyMail and Gigaword after pre-training, where the setting is the same as described in \S~\ref{sec:exp2}. 
We compare ProphetNet (160GB) against the following strong baselines: \textbf{T5}~\cite{raffel2019exploring} which is pre-trained on the text corpus of 750GB; \textbf{PEGASUS}\textsubscript{LARGE}~\cite{zhang2019pegasus} which is pre-trained on the text corpus of 750GB and 3800GB, respectively; And \textbf{BART}~\cite{lewis2019bart} which is pre-trained on the similar dataset as the ProphetNet (160GB).

We pre-train our model on 16 $\times$ 32GB NVIDIA V100 GPUs with 14 epochs.  We can see that the performance increase as ProphetNet pre-trains for more epochs on 160GB large-scale dataset.  The results on test set are shown in Table~\ref{tab:large}.   
Our model achieves state-of-the-art performance on CNN/DailyMail compared to other baselines.
It can be observed that the ROUGE-1 and ROUGE-L of ProphetNet on CNN/DailyMail are the highest.
Moreover, ProphetNet (160GB) outperforms PEGASUS\textsubscript{LARGE} (C4 750GB) and PEGASUS\textsubscript{LARGE} (HugeNews 3800GB) on Gigaword using only about 1/5 and 1/20 of the pre-training corpus, respectively. To the best of our knowledge, ProphetNet also achieves new state-of-the-art results on the Gigaword. 

\subsection{ProphetNet without Pre-training}~\label{sec:exp5}
ProphetNet achieves significant results improvement after pre-training, we also curious about the performance of ProphetNet when directly applied it to downstream tasks without pre-training.
Therefore, we evaluate the ProphetNet model without pre-training on CNN/DailyMail.
The ProphetNet model without pre-training consists of 12-layer encoder and 12-layer decoder with 768 embedding/hidden size and 3072 feed-forward filter size.
We compare the ProphetNet model with the original Seq2Seq Transformer which has the same architecture hyper-parameters of the ProphetNet.
The training and evaluation details are the same as described in \S~\ref{sec:exp2}.
The results are shown in Table~\ref{tab:nopretrain}.
Experimental results show that our method can significantly improve the model performance even without pre-training.
\begin{table}[h] 
\small
\begin{center}
\begin{tabular}{lcccccl}
\toprule
    Setting & R-1 & R-2 & R-L \\
    \midrule
 Transformer\scriptsize{~\cite{raffel2019exploring}} & 39.19    &17.60&   36.69\\
 ProphetNet\textsubscript{w/o pre-train} & \textbf{40.66}    & \textbf{18.05} & \textbf{37.79}\\
  \bottomrule
\end{tabular}
\end{center}
\caption{Results on CNN/DailyMail dev set without pre-training}
\label{tab:nopretrain}
\end{table}

\subsection{ProphetNet N-gram Comparison}~\label{sec:exp6}
ProphetNet predicts next contiguous $n$-gram tokens simultaneously for each time step. To explore the effectiveness of predicting $n$ gram, we compare our ProphetNet model with $n$=1, 2, and 3. 
We also compare the MASS\textsubscript{base} which is very similar to ProphetNet\textsubscript{base}-1gram.  The architecture hyper-parameter of all the models is set to 6-layer encoder, 6-layer decoder, 768 hidden size, and 12 attention heads, which are the same as MASS\textsubscript{base}. These models are also pre-trained on the Wikipedia+BookCorpus dataset with 125k steps. Other hyper-parameters are the same as the description in \S~\ref{sec:exp1}. As we mentioned in \S~\ref{sec:m1}, we set different attenuation coefficient for the power attenuation function. For ProphetNet\textsubscript{base}-2gram, $\gamma$ is set to 1.0. For ProphetNet\textsubscript{base}-3gram model, the attenuation coefficient $\gamma$ is set to 0.5.

The pre-trained models are then fine-tuned on CNN/DailyMail.
We report the F1 scores of ROUGE-1, ROUGE-2 and ROUGE-L. The results are shown in Table~\ref{tab:abngram}. We can see that the performance of ProphetNet\textsubscript{base}-3gram and ProphetNet\textsubscript{base}-2gram is comparable. Both of them perform better than MASS\textsubscript{base} and ProphetNet\textsubscript{base}-1gram. Considering the computational and time cost, we use ProphetNet\textsubscript{base}-2gram in other experiments due to its training speed is 15\% faster than ProphetNet\textsubscript{base}-3gram.

\begin{table}[ht] 
\small
\begin{center}
  \begin{tabular}{lcccccl}
    \toprule
    Setting & R-1 & R-2 & R-L \\
    \midrule
 MASS\textsubscript{base}& 42.12 &19.50&   39.01\\
  ProphetNet\textsubscript{base}-1gram &   42.21  & 19.54   &39.06\\
 ProphetNet\textsubscript{base}-2gram &   42.52  & 19.78   &39.59\\
 ProphetNet\textsubscript{base}-3gram & \textbf{42.61}    & \textbf{19.83} &   \textbf{39.67}\\
  \bottomrule
\end{tabular}
\end{center}
\caption{n-gram comparison results on CNN/DailyMail test set}
\label{tab:abngram}
\end{table}

\section{Related Work}
Unsupervised pre-training has been successfully applied to various natural language processing tasks. 
GPT~\cite{radford2018improving} takes plain text as pre-training data to predict the next tokens with leftward tokens. It is based on the left-to-right language model and can be used to generate stories and continue to write for a given text. 
BERT~\cite{devlin2018bert} and SpanBERT~\cite{joshi2019spanbert} use a Bi-directional language model to recover masked tokens/spans for a given sentence. Bi-directional information flow can be used to recover the masked positions, but no left-to-right language model dependency is learned. As a result, BERT and SpanBERT bring significant improvement for NLU tasks but are not suitable for generation tasks. XLNet~\cite{yang2019xlnet} predicts the tokens with given positions and some tokens with their positions in the sentence in an AR manner. Although it uses AR to build a permuted-ordered language model, it is also not suitable for NLG tasks because it brought too much noise for a left-to-right language model. 
MASS~\cite{song2019mass} pre-trains the sequence-to-sequence model by dropping a continuous token span to corrupt the original text and learns to recover it. 
T5~\cite{raffel2019exploring} investigates different model structures and different pre-training tasks, and is pre-trained on a large scale corpus named C4 which is 750GB.
BART~\cite{lewis2019bart} uses the encoder-decoder structure to generate the original sentence with its spoiled input to denoise. In the BART decoder, the undamaged language model is learned thus brings improvement to NLG tasks. 

Natural language generation methods are typically based on the left-to-right or right-to-left language models and generate one token in each time step. These methods can not capture the information of future tokens. Recently, incorporating future information into language generation tasks has attracted the attention of researchers~\cite{li2017learning,serdyuk2017twin,lawrence2019attending,oord2018representation}.~\citet{li2017learning} propose an actor-critic model which designs a value function as a critic to estimate the future success. In their method, they not only consider the MLE-based learning but also incorporate an RL-based value function into the decoder process.~\cite{oord2018representation} do not predict future tokens directly but tried to model a density ratio to preserve the mutual information between context and future token.~\citet{serdyuk2017twin} point out traditional Recurrent Neural Networks (RNNs) may prefer to generate each token based on the recent tokens, it is hard to learn the long-term dependencies. To capture the future information and learn the long-term dependencies, they run the forward RNN and backward RNN in parallel.~\citet{lawrence2019attending} concatenates the source and target to train an encoder instead of encoder-decoder architecture. They use special placeholder tokens to replace some tokens of the target for the model training process. At the inference process, they generate the target by replacing each placeholder token.

\section{Conclusion}
In this paper, we introduce ProphetNet, a sequence-to-sequence pre-training model that learns to predict future n-gram at each time step.
ProphetNet achieves the best performance on both abstractive summarization and question generation tasks. Furthermore, ProphetNet achieves new state-of-the-art results on CNN/DailyMail and Gigaword using only about 1/3 the pre-training epochs of the previous model. 


\newpage

\bibliographystyle{acl_natbib}
\bibliography{emnlp2020}

\begin{thebibliography}{41}
\expandafter\ifx\csname natexlab\endcsname\relax\def\natexlab#1{#1}\fi

\bibitem[{Cao et~al.(2018)Cao, Li, Li, and Wei}]{cao2018retrieve}
Ziqiang Cao, Wenjie Li, Sujian Li, and Furu Wei. 2018.
\newblock Retrieve, rerank and rewrite: Soft template based neural
  summarization.
\newblock In \emph{ACL}.

\bibitem[{Devlin et~al.(2018)Devlin, Chang, Lee, and
  Toutanova}]{devlin2018bert}
Jacob Devlin, Ming-Wei Chang, Kenton Lee, and Kristina Toutanova. 2018.
\newblock Bert: Pre-training of deep bidirectional transformers for language
  understanding.
\newblock In \emph{NAACL}.

\bibitem[{Dong et~al.(2019)Dong, Yang, Wang, Wei, Liu, Wang, Gao, Zhou, and
  Hon}]{dong2019unified}
Li~Dong, Nan Yang, Wenhui Wang, Furu Wei, Xiaodong Liu, Yu~Wang, Jianfeng Gao,
  Ming Zhou, and Hsiao-Wuen Hon. 2019.
\newblock Unified language model pre-training for natural language
  understanding and generation.
\newblock In \emph{NeurIPS}.

\bibitem[{Du and Cardie(2018)}]{du2018harvesting}
Xinya Du and Claire Cardie. 2018.
\newblock Harvesting paragraph-level question-answer pairs from wikipedia.
\newblock In \emph{ACL}.

\bibitem[{Du et~al.(2017)Du, Shao, and Cardie}]{du2017learning}
Xinya Du, Junru Shao, and Claire Cardie. 2017.
\newblock Learning to ask: Neural question generation for reading
  comprehension.
\newblock \emph{arXiv preprint arXiv:1705.00106}.

\bibitem[{Edunov et~al.(2019)Edunov, Baevski, and Auli}]{edunov2019pre}
Sergey Edunov, Alexei Baevski, and Michael Auli. 2019.
\newblock Pre-trained language model representations for language generation.
\newblock \emph{arXiv preprint arXiv:1903.09722}.

\bibitem[{Fan et~al.(2017)Fan, Grangier, and Auli}]{fan2017controllable}
Angela Fan, David Grangier, and Michael Auli. 2017.
\newblock Controllable abstractive summarization.
\newblock \emph{arXiv preprint arXiv:1711.05217}.

\bibitem[{Gehrmann et~al.(2018)Gehrmann, Deng, and Rush}]{gehrmann2018bottom}
Sebastian Gehrmann, Yuntian Deng, and Alexander~M Rush. 2018.
\newblock Bottom-up abstractive summarization.
\newblock In \emph{EMNLP}.

\bibitem[{Gulcehre et~al.(2017)Gulcehre, Dutil, Trischler, and
  Bengio}]{gulcehre2017plan}
Caglar Gulcehre, Francis Dutil, Adam Trischler, and Yoshua Bengio. 2017.
\newblock Plan, attend, generate: Planning for sequence-to-sequence models.
\newblock In \emph{NIPS}.

\bibitem[{Joshi et~al.(2019)Joshi, Chen, Liu, Weld, Zettlemoyer, and
  Levy}]{joshi2019spanbert}
Mandar Joshi, Danqi Chen, Yinhan Liu, Daniel~S Weld, Luke Zettlemoyer, and Omer
  Levy. 2019.
\newblock Spanbert: Improving pre-training by representing and predicting
  spans.
\newblock \emph{arXiv preprint arXiv:1907.10529}.

\bibitem[{Kingma and Ba(2015)}]{kingma2014adam}
Diederik~P Kingma and Jimmy Ba. 2015.
\newblock Adam: A method for stochastic optimization.
\newblock In \emph{ICLR}.

\bibitem[{Klein et~al.(2017)Klein, Kim, Deng, Senellart, and
  Rush}]{klein2017opennmt}
Guillaume Klein, Yoon Kim, Yuntian Deng, Jean Senellart, and Alexander~M Rush.
  2017.
\newblock Opennmt: Open-source toolkit for neural machine translation.
\newblock In \emph{ACL}.

\bibitem[{Krueger et~al.(2016)Krueger, Maharaj, Kram{\'a}r, Pezeshki, Ballas,
  Ke, Goyal, Bengio, Courville, and Pal}]{krueger2016zoneout}
David Krueger, Tegan Maharaj, J{\'a}nos Kram{\'a}r, Mohammad Pezeshki, Nicolas
  Ballas, Nan~Rosemary Ke, Anirudh Goyal, Yoshua Bengio, Aaron Courville, and
  Chris Pal. 2016.
\newblock Zoneout: Regularizing rnns by randomly preserving hidden activations.
\newblock \emph{arXiv preprint arXiv:1606.01305}.

\bibitem[{Lawrence et~al.(2019)Lawrence, Kotnis, and
  Niepert}]{lawrence2019attending}
Carolin Lawrence, Bhushan Kotnis, and Mathias Niepert. 2019.
\newblock Attending to future tokens for bidirectional sequence generation.
\newblock \emph{arXiv preprint arXiv:1908.05915}.

\bibitem[{Lewis et~al.(2019)Lewis, Liu, Goyal, Ghazvininejad, Mohamed, Levy,
  Stoyanov, and Zettlemoyer}]{lewis2019bart}
Mike Lewis, Yinhan Liu, Naman Goyal, Marjan Ghazvininejad, Abdelrahman Mohamed,
  Omer Levy, Ves Stoyanov, and Luke Zettlemoyer. 2019.
\newblock Bart: Denoising sequence-to-sequence pre-training for natural
  language generation, translation, and comprehension.
\newblock \emph{arXiv preprint arXiv:1910.13461}.

\bibitem[{Li et~al.(2017)Li, Monroe, and Jurafsky}]{li2017learning}
Jiwei Li, Will Monroe, and Dan Jurafsky. 2017.
\newblock Learning to decode for future success.
\newblock \emph{arXiv preprint arXiv:1701.06549}.

\bibitem[{Lin(2004)}]{lin2004rouge}
Chin-Yew Lin. 2004.
\newblock Rouge: A package for automatic evaluation of summaries.
\newblock In \emph{Text summarization branches out}.

\bibitem[{Liu and Lapata(2019)}]{liu2019text}
Yang Liu and Mirella Lapata. 2019.
\newblock Text summarization with pretrained encoders.
\newblock \emph{arXiv preprint arXiv:1908.08345}.

\bibitem[{Merity et~al.(2017)Merity, Keskar, and
  Socher}]{merity2017regularizing}
Stephen Merity, Nitish~Shirish Keskar, and Richard Socher. 2017.
\newblock Regularizing and optimizing lstm language models.
\newblock \emph{arXiv preprint arXiv:1708.02182}.

\bibitem[{Nallapati et~al.(2017)Nallapati, Zhai, and
  Zhou}]{nallapati2017summarunner}
Ramesh Nallapati, Feifei Zhai, and Bowen Zhou. 2017.
\newblock Summarunner: A recurrent neural network based sequence model for
  extractive summarization of documents.
\newblock In \emph{AAAI}.

\bibitem[{Nallapati et~al.(2016)Nallapati, Zhou, Gulcehre, Xiang
  et~al.}]{nallapati2016abstractive}
Ramesh Nallapati, Bowen Zhou, Caglar Gulcehre, Bing Xiang, et~al. 2016.
\newblock Abstractive text summarization using sequence-to-sequence rnns and
  beyond.
\newblock \emph{arXiv preprint arXiv:1602.06023}.

\bibitem[{Oord et~al.(2018)Oord, Li, and Vinyals}]{oord2018representation}
Aaron van~den Oord, Yazhe Li, and Oriol Vinyals. 2018.
\newblock Representation learning with contrastive predictive coding.
\newblock \emph{arXiv preprint arXiv:1807.03748}.

\bibitem[{Pascanu et~al.(2013)Pascanu, Mikolov, and
  Bengio}]{pascanu2013difficulty}
Razvan Pascanu, Tomas Mikolov, and Yoshua Bengio. 2013.
\newblock On the difficulty of training recurrent neural networks.
\newblock In \emph{ICML}.

\bibitem[{Peters et~al.(2018)Peters, Neumann, Iyyer, Gardner, Clark, Lee, and
  Zettlemoyer}]{peters2018deep}
Matthew~E Peters, Mark Neumann, Mohit Iyyer, Matt Gardner, Christopher Clark,
  Kenton Lee, and Luke Zettlemoyer. 2018.
\newblock Deep contextualized word representations.
\newblock In \emph{NAACL}.

\bibitem[{Radford et~al.(2018)Radford, Narasimhan, Salimans, and
  Sutskever}]{radford2018improving}
Alec Radford, Karthik Narasimhan, Tim Salimans, and Ilya Sutskever. 2018.
\newblock Improving language understanding by generative pre-training.
\newblock \emph{URL https://s3-us-west-2. amazonaws.
  com/openai-assets/research-covers/languageunsupervised/language understanding
  paper. pdf}.

\bibitem[{Radford et~al.(2019)Radford, Wu, Child, Luan, Amodei, and
  Sutskever}]{radford2019language}
Alec Radford, Jeffrey Wu, Rewon Child, David Luan, Dario Amodei, and Ilya
  Sutskever. 2019.
\newblock Language models are unsupervised multitask learners.
\newblock \emph{OpenAI Blog}, 1(8).

\bibitem[{Raffel et~al.(2019)Raffel, Shazeer, Roberts, Lee, Narang, Matena,
  Zhou, Li, and Liu}]{raffel2019exploring}
Colin Raffel, Noam Shazeer, Adam Roberts, Katherine Lee, Sharan Narang, Michael
  Matena, Yanqi Zhou, Wei Li, and Peter~J Liu. 2019.
\newblock Exploring the limits of transfer learning with a unified text-to-text
  transformer.
\newblock \emph{arXiv preprint arXiv:1910.10683}.

\bibitem[{Rajpurkar et~al.(2016)Rajpurkar, Zhang, Lopyrev, and
  Liang}]{rajpurkar2016squad}
Pranav Rajpurkar, Jian Zhang, Konstantin Lopyrev, and Percy Liang. 2016.
\newblock Squad: 100,000+ questions for machine comprehension of text.
\newblock In \emph{EMNLP}.

\bibitem[{Rush et~al.(2015)Rush, Chopra, and Weston}]{rush2015neural}
Alexander~M Rush, Sumit Chopra, and Jason Weston. 2015.
\newblock A neural attention model for abstractive sentence summarization.
\newblock \emph{arXiv preprint arXiv:1509.00685}.

\bibitem[{See et~al.(2017)See, Liu, and Manning}]{see2017get}
Abigail See, Peter~J Liu, and Christopher~D Manning. 2017.
\newblock Get to the point: Summarization with pointer-generator networks.
\newblock In \emph{ACL}.

\bibitem[{Serdyuk et~al.(2018)Serdyuk, Ke, Sordoni, Trischler, Pal, and
  Bengio}]{serdyuk2017twin}
Dmitriy Serdyuk, Nan~Rosemary Ke, Alessandro Sordoni, Adam Trischler, Chris
  Pal, and Yoshua Bengio. 2018.
\newblock Twin networks: Matching the future for sequence generation.
\newblock In \emph{ICLR}.

\bibitem[{Song et~al.(2019)Song, Tan, Qin, Lu, and Liu}]{song2019mass}
Kaitao Song, Xu~Tan, Tao Qin, Jianfeng Lu, and Tie-Yan Liu. 2019.
\newblock Mass: Masked sequence to sequence pre-training for language
  generation.
\newblock \emph{arXiv preprint arXiv:1905.02450}.

\bibitem[{Sutskever et~al.(2014)Sutskever, Vinyals, and
  Le}]{sutskever2014sequence}
Ilya Sutskever, Oriol Vinyals, and Quoc~V Le. 2014.
\newblock Sequence to sequence learning with neural networks.
\newblock In \emph{NIPS}.

\bibitem[{Vaswani et~al.(2017)Vaswani, Shazeer, Parmar, Uszkoreit, Jones,
  Gomez, Kaiser, and Polosukhin}]{vaswani2017attention}
Ashish Vaswani, Noam Shazeer, Niki Parmar, Jakob Uszkoreit, Llion Jones,
  Aidan~N Gomez, {\L}ukasz Kaiser, and Illia Polosukhin. 2017.
\newblock Attention is all you need.
\newblock In \emph{NIPS}.

\bibitem[{Yang et~al.(2019)Yang, Dai, Yang, Carbonell, Salakhutdinov, and
  Le}]{yang2019xlnet}
Zhilin Yang, Zihang Dai, Yiming Yang, Jaime Carbonell, Ruslan Salakhutdinov,
  and Quoc~V Le. 2019.
\newblock Xlnet: Generalized autoregressive pretraining for language
  understanding.
\newblock \emph{arXiv preprint arXiv:1906.08237}.

\bibitem[{Zellers et~al.(2019)Zellers, Holtzman, Rashkin, Bisk, Farhadi,
  Roesner, and Choi}]{zellers2019defending}
Rowan Zellers, Ari Holtzman, Hannah Rashkin, Yonatan Bisk, Ali Farhadi,
  Franziska Roesner, and Yejin Choi. 2019.
\newblock Defending against neural fake news.
\newblock \emph{arXiv preprint arXiv:1905.12616}.

\bibitem[{Zhang et~al.(2019)Zhang, Zhao, Saleh, and Liu}]{zhang2019pegasus}
Jingqing Zhang, Yao Zhao, Mohammad Saleh, and Peter~J Liu. 2019.
\newblock Pegasus: Pre-training with extracted gap-sentences for abstractive
  summarization.
\newblock \emph{arXiv preprint arXiv:1912.08777}.

\bibitem[{Zhang and Bansal(2019)}]{zhang2019addressing}
Shiyue Zhang and Mohit Bansal. 2019.
\newblock Addressing semantic drift in question generation for semi-supervised
  question answering.
\newblock \emph{arXiv preprint arXiv:1909.06356}.

\bibitem[{Zhao et~al.(2018)Zhao, Ni, Ding, and Ke}]{zhao2018paragraph}
Yao Zhao, Xiaochuan Ni, Yuanyuan Ding, and Qifa Ke. 2018.
\newblock Paragraph-level neural question generation with maxout pointer and
  gated self-attention networks.
\newblock In \emph{EMNLP}.

\bibitem[{Zhou et~al.(2017)Zhou, Yang, Wei, Tan, Bao, and
  Zhou}]{zhou2017neural}
Qingyu Zhou, Nan Yang, Furu Wei, Chuanqi Tan, Hangbo Bao, and Ming Zhou. 2017.
\newblock Neural question generation from text: A preliminary study.
\newblock In \emph{National CCF Conference on Natural Language Processing and
  Chinese Computing}, pages 662--671.

\bibitem[{Zhu et~al.(2015)Zhu, Kiros, Zemel, Salakhutdinov, Urtasun, Torralba,
  and Fidler}]{zhu2015aligning}
Yukun Zhu, Ryan Kiros, Rich Zemel, Ruslan Salakhutdinov, Raquel Urtasun,
  Antonio Torralba, and Sanja Fidler. 2015.
\newblock Aligning books and movies: Towards story-like visual explanations by
  watching movies and reading books.
\newblock In \emph{Proceedings of the IEEE international conference on computer
  vision}, pages 19--27.

\end{thebibliography}

\end{document}